\documentclass[sn-mathphys-num]{sn-jnl}% Math and Physical Sciences Numbered Reference Style 
\usepackage{graphicx}%
\usepackage{multirow}%
\usepackage{amsmath,amssymb,amsfonts}%
\usepackage{amsthm}%
\usepackage{mathrsfs}%
\usepackage[title]{appendix}%
\usepackage{xcolor}%
\usepackage{textcomp}%
\usepackage{manyfoot}%
\usepackage{booktabs}%
\usepackage{algorithm}%
\usepackage{algorithmicx}%
\usepackage{algpseudocode}%
\usepackage{listings}%
\usepackage{caption}

\theoremstyle{thmstyleone}%
\theoremstyle{thmstyletwo}%

\theoremstyle{thmstylethree}%

\begin{document}

\title[RoHan]{RoHan: Robust Hand Detection in Operation
Room }


\author*[1]{\fnm{Roi} \sur{Papo}}\email{roipapo@campus.technion.ac.il}

\author[1]{\fnm{Sapir} \sur{Gershov}}

\author[2]{Tom \sur{Friedman}}

\author[2]{Itay \sur{Or}}

\author[2]{Gil \sur{Bolotin}}

\author[1]{\fnm{Shlomi} \sur{Laufer}}

\affil*[1]{\orgdiv{Faculty of Data and Decision Sciences}, \orgname{Technion – Israel Institute of
Technology}, \orgaddress{\city{Haifa}, \postcode{3200003}, \country{Israel}}}
\affil*[2]{ \orgname{Rambam Health Care Campus},  \orgaddress{\city{Haifa}, \country{Israel}}}


\abstract{Hand-specific localization has garnered significant interest within the computer vision community. Although there are numerous datasets with hand annotations from various angles and settings, domain transfer techniques frequently struggle in surgical environments. This is mainly due to the limited availability of gloved hand instances and the unique challenges of operating rooms (ORs). Thus, hand-detection models tailored to OR settings require extensive training and expensive annotation processes. To overcome these challenges, we present "RoHan" - a novel approach for robust hand detection in the OR, leveraging advanced semi-supervised domain adaptation techniques to tackle the challenges of varying recording conditions, diverse glove colors, and occlusions common in surgical settings. Our methodology encompasses two main stages: (1) data augmentation strategy that utilizes "Artificial Gloves," a method for augmenting publicly available hand datasets with synthetic images of hands-wearing gloves; (2)  semi-supervised domain adaptation pipeline that improves detection performance in real-world OR settings through iterative prediction refinement and efficient frame filtering.
We evaluate our method using two datasets: simulated enterotomy repair and saphenous vein graft harvesting. "RoHan" substantially reduces the need for extensive labeling and model training, paving the way for the practical implementation of hand detection technologies in medical settings.
Code available at: \url{https://github.com/RoiPapo/Rohan}
}




\maketitle
\begin{figure}[!htb]
    \centering
    \includegraphics[scale=0.4]{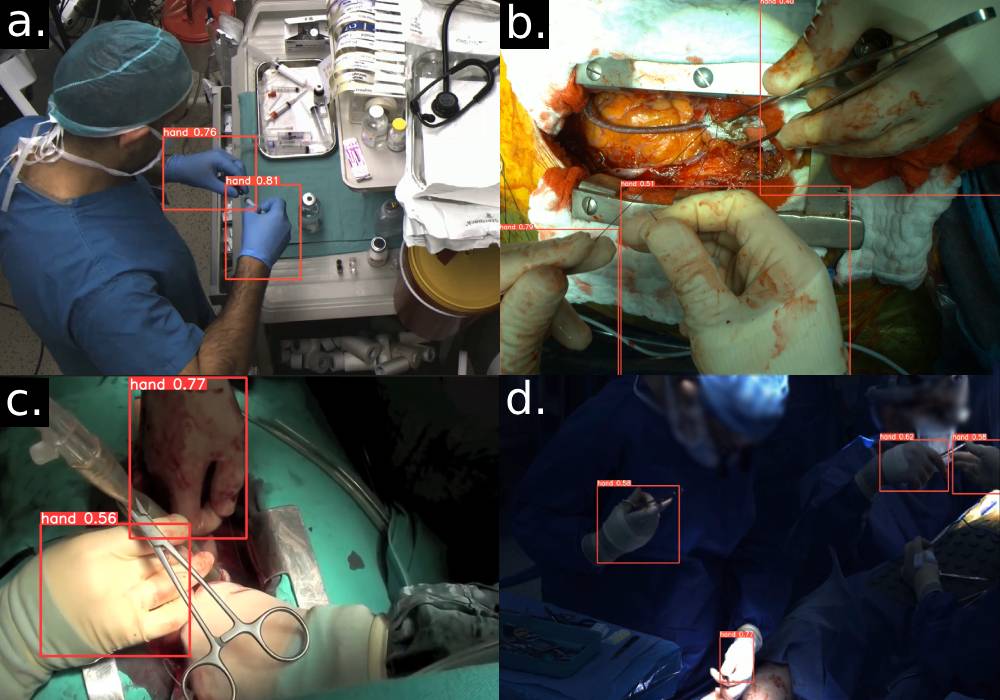}
    \caption{\textbf{Robust Hand Detection} We propose the use of Semi-Supervised Domain Adaptation and Self-Training Techniques for hand detection. These techniques provide predictions that demonstrate robustness across various shooting angles and lighting conditions. Specifically, they are effective in scenarios including: a) top views, b) views from a surgeon's head-mounted camera, c) zoomed-in views of the surgeon's working area, and d) side views of the patient's bed. Faces were manually blured} 
\label{fig robustness}
\end{figure}

\section{Introduction}\label{sec1}

The precision of hand movements in surgical procedures is paramount, directly correlating with the reduction of errors and the efficiency of operations, thereby ensuring the execution of precise and secure surgeries \cite{kim2015,huang2011}. The advent of computer vision algorithms capable of autonomously detecting surgeon hand movements within the operating room (OR) heralds a new era of potential enhancements in surgical practices \cite{ward2021}. Furthermore, the ability of hand detection models to objectively evaluate surgical techniques presents an opportunity to revolutionize the assessment process, traditionally dependent on the subjective, costly, and time-consuming visual inspections by human experts \cite{zhang2020}.

However, the development of effective hand detection models is constrained by significant challenges. The need for extensive labeled data and rigorous training collides with the privacy concerns within the OR setting. Although transfer learning has been utilized as a standard mitigating strategy, the lack of publicly available datasets featuring gloved hands limits the accuracy of models in recognizing surgeons' hands \cite{vaid2023}.
Another challenge is posed by the diverse range of glove colors and textures used in ORs, which complicates the ability of domain-specific hand detection models to recognize surgeons' hands accurately. The variability in glove appearances, compounded by the potential obscuring effects of blood and the inconsistency of lighting conditions and camera angles in ORs, further exacerbates the difficulty in achieving reliable hand detection \cite{chen2019}.

To address the degradation in model performance when transitioning from general hand datasets to surgical data, we introduce 'Artificial Gloves' - a novel augmentation technique that overlays synthetic gloves onto images of bare hands. We then incorporate an innovative iterative prediction refinement process to further enhance the model's performance. First, we pre-train a detection model on the augmented datasets (i.e., Artificial Gloves). Then, using the temporal characteristics of video data, we refine the pseudo-labels to reduce inaccuracies. Lastly, the detection model is fine-tuned once more using the cleaned pseudo dataset, which significantly improves the model's robustness and accuracy across diverse medical settings. Finally, we test our approach using two datasets: an enterotomy repair simulator (ERS) \cite{basiev2022open} and a new dataset of saphenous vein graft harvesting (SVGH) during coronary artery bypass graft surgery. We utilized the YOLO framework \cite{jiang2022}, a family of real-time object detection models, to explore its adaptability across different hand datasets. 

This research contribution is threefold: first, a new dataset is established. Second we propose a unique solution to the challenges of hand detection in surgical environments through the use of 'Artificial Gloves,' a domain-specific augmentation. Last, our iterative prediction refinement method offers a scalable, efficient approach to enhancing model performance without requiring extensive new data labeling or extensive training. Hence, this work paves the way for advanced hand detection models tailored to the specific needs of OR environments.

\section{Related Work}
The emergence of computer vision and deep learning techniques has revolutionized the automation and enhancement of various medical process aspects, notably in detecting and tracking medical staff hand movements in operating rooms (ORs). Zhang et al. \cite{zhang2020} delved into hand detection by annotating hands in publicly accessible open surgery videos from YouTube with spatial bounding boxes. Employing a Convolutional Neural Network (CNN) architecture, their approach markedly improved detection performance compared to traditional hand-detection datasets. Goodman et al. \cite{goodman2021} implemented a hand detection system leveraging the RetinaNet architecture, integrated with a shared Feature Pyramid Network (FPN) for detecting objects across multiple scales. This system was trained on the Annotated Videos of Open Surgery (AVOS) Dataset. Though showing promising results, both studies highly depend on large amounts of human-expert labeled data.

Recently, Vaid et al. \cite{vaid2023} proposed a semi-supervised two-stage hand detection framework that is end-to-end trainable and incorporates a consistency loss to facilitate learning from unlabeled images. Despite these innovations, their approach did not surpass the performance of fully supervised methods on the AVOS hands dataset. The diverse and complex nature of surgical data in the real world hampered the semi-supervised model's learning efficiency, reducing accuracy. 

The works mentioned above have all utilized the AVOS dataset \cite{goodman2021} as an integral part of their training procedures. This dataset includes 3,430 frames, with every ten frames uniformly sampled from each video. Each frame is densely annotated with spatial labels for hands and tools. However, our methodology relied on temporal information, which is limited in this dataset. Therefore, we evaluate the robustness of our methods on the ERS and SVGH datasets.

Recent advancements in 3D bare hand reconstruction using transformer-based approaches have shown promise in improving robustness and accuracy across varied conditions \cite{pavlakos2024}. These methods present strong detectors as a fundamental puzzle piece and leverage large-scale datasets and high-capacity architectures to enhance performance even in challenging environments. Additionally, novel pipelines for real-time hand localization and reconstruction have demonstrated state-of-the-art reconstructing results by addressing limitations in existing detection frameworks through refinement modules and large-scale data collection. \cite{potamias2024}.The detection components of the previously discussed models were evaluated using ERS and
SVGH datasets and demonstrated difficulties in accurately identifying gloved hands.

\section{Methods}
\subsection{Datasets Used for Pre-training}
For the pre-trained hand-detection model, we utilized the following group of datasets:

\noindent \textbf{EgoHands} \cite{bambach2015} includes 4,800 images that capture intricate, first-person interactions between two individuals and are labeled with binary segmentation masks of the hands. The masked were utilized to create artificial gloves, and the mask's bounding box for detection labels.

\noindent \textbf{OneHand10K} 
\cite{wang2018mask} comprises $10,000$ labeled images featuring a single hand in diverse scenarios and lighting conditions. The hands are prominently displayed directly in front of the camera, a perspective not typical in surgical settings. Once again the binary masks were used for generating gloves and bounding boxes.

\noindent \textbf{HaDR} \cite{grushko2023} is a multi-modal dataset tailored for human-robot gesture-based interaction. It encompasses $108,650$ frames and binary masks for each hand instance. Notably, the dataset is entirely synthetic, generated through Domain Randomization techniques. 
The integration of Artificial Gloves was not needed for this dataset as this dataset presented hands with various colors.

\noindent \textbf{Assembly 101} \cite{sener2022assembly101} features multi-camera data of toy assembly without segmentation masks. For segmentation masks, We randomly selected 8,000 frames and applied the InterWild model \cite{moon2023bringing} to simulate 3D artificial gloves. However, the InterWild results were unsatisfactory. Consequently, we manually chose 1,000 frames with better outcomes. Given the smaller scale of this dataset, it was unsuitable for training a detection model on its own. Instead, these augmented frames were incorporated into a larger combined dataset, selected for its camera angles that resemble those in operating rooms.

\subsection{Datasets Used for Benchmark}
We used the following datasets to establish a benchmark to evaluate the pre-trained hand detectors and the iterative pipeline.

\noindent \textbf{Enterotomy Repair Simulator (ERS)}: 
Attending and retired surgeons were requested to conduct a simulated repair of a bowel enterotomy \cite{basiev2022open}. Data was captured from two angles. From seven randomly chosen simulations, $2407$ frames were extracted, and hands' bounding boxes were labeled. The surgeons used white sterile gloves, whereas, in contrast to actual surgery, the surgical assistant wore basic non-sterile blue gloves. Medical personnel often use these gloves in the OR  while performing non-sterile tasks, hence the value of identifying them. We captured each frame from two angles - the top view and the surface view.

\begin{figure}[!htb]
    \centering
    \includegraphics[scale=0.1]{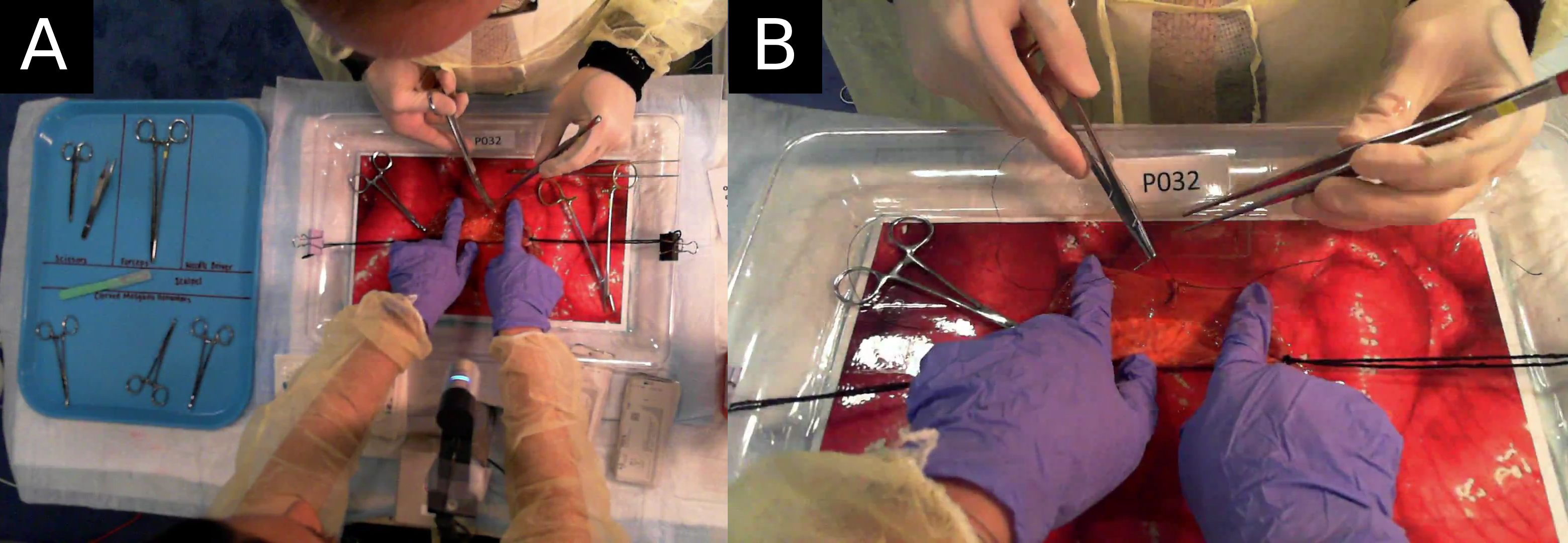}
    \caption{Frames from the ERS dataset. (A) Top view; (B) Surface view.} 
    \label{fig:1}
\end{figure}

\noindent \textbf{Saphenous Vein Graft Harvesting (SVGH)}:
Four Saphenous Vein Graft Harvesting procedures, executed by two residents and one attending surgeon, were documented. A Back-Bone GoPro H7PRO camera (https://www.back-bone.ca/) was mounted on a tripod about 1 meter from the leg. The Back-Bone model supports the use of various lenses. Four one-minute video segments were extracted from each procedure. During the refinement stage, these videos were sampled at five fps. In addition, the hands in 480 frames were annotated.

\subsection{Artificial Gloves}
\begin{figure}[!htb]
    \centering
    \includegraphics[scale=0.185]{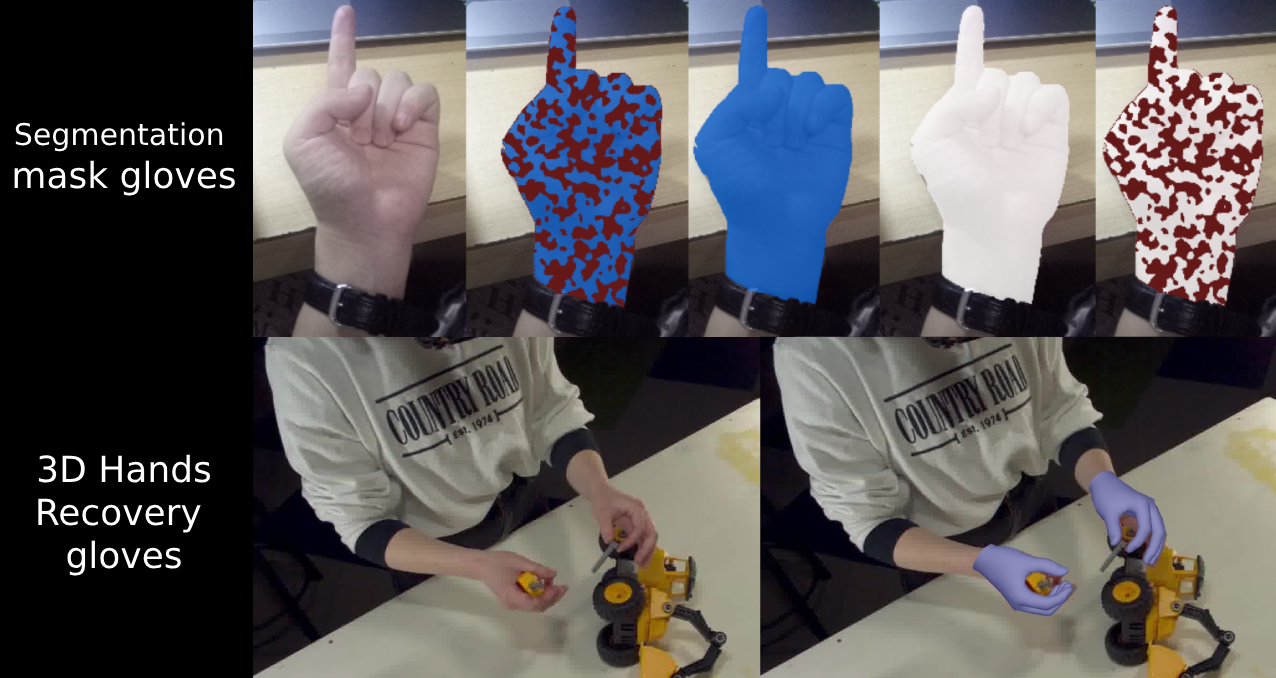}
   \caption{In the images above, Artificial Gloves are applied to public datasets using the previously mentioned Segmentation mask gloves method, featuring various glove colors and blood patterns. In the image below, 3D Hands Recovery gloves are utilized to simulate gloves.}
   \label{fig:2}
\end{figure}

An innovative technique was developed to create "Segmentation mask gloves". A low-opacity, glove-colored mask was applied to the original hand based on its segmentation mask. The result was a realistic overlay of semi-transparent artificial gloves on the hand. For the Assembly 101 dataset, the InterWild model \cite{moon2023bringing} was employed to generate "3D Hands Recovery gloves" (see Fig. \ref{fig:2}).
This process yields a synthetic representation of hands wearing artificial gloves, bridging the gap between surgical and non-surgical hand appearances. In operating rooms, gloves are typically white or blue and may be covered in blood—a feature not present in public, non-medical hand datasets. To simulate blood on the gloves, random noise was generated. Using blurring effects, and utilizing morphological operations the noise was refined to splatter shapes, resulting in realistic blood patterns on the synthetic gloves.
By incorporating these artificial gloves and simulated blood patterns, the augmentation technique helps to mitigate the visual discrepancies between surgical and non-surgical hands, enhancing the model's ability to generalize across various hand appearances and contexts.

\subsection{Domain Adaptation Pipeline}
%To enhance our ability to accurately and consistently identify hands in the OR, we addressed the discrepancies in the distribution between training data, which lacks representation of gloves, and testing data, which consists of gloved hands. 
To mitigate the discrepancies in the distribution between the training datasets and the target data, we introduce a novel fine-tuning domain adaptation pipeline.
This technique aims to improve the model's performance and adaptability across different datasets. The process is iterative, with the output of each iteration providing pseudo-labels for the next round, as outlined below.  

\noindent \textbf{Data Preprocessing.}
To initialize the pipeline, unseen videos from the new domain are processed using the hand detection model trained on the existing datasets.
This step generates an initial pseudo-dataset, in which the original frames are paired with model predictions serving as pseudo-labels.

\noindent \textbf{Pseudo Labels Enhancement(Auto Filtering Best Frames).}
Using the temporal properties of video, a filtering mechanism is subsequently applied to the initial pseudo-dataset to eliminate frames of inferior prediction quality. The output of this stage consists of refined pseudo labels. Two filtering mechanisms were assessed and are described below.

\noindent \textbf{Model Fine-Tuning.} 
The filtered pseudo-dataset is used to fine-tune the Yolo model to the unique characteristics of the unseen environmental conditions. In our experiments, we fine-tune the model for five epochs. The steps of Data Preprocessing to Model Fine-Tuning are repeated to refine the model further, creating an iterative loop that refines the model weights.

\noindent \textbf{Model Evaluation.}
Finally, the refined model weights are assessed by testing them on the target test data. This evaluation step gauges the adaptability and generalization capability of the fine-tuned model in the face of previously unencountered environmental conditions.

\begin{figure}[!htb]
    \centering
    \includegraphics[width=\textwidth]{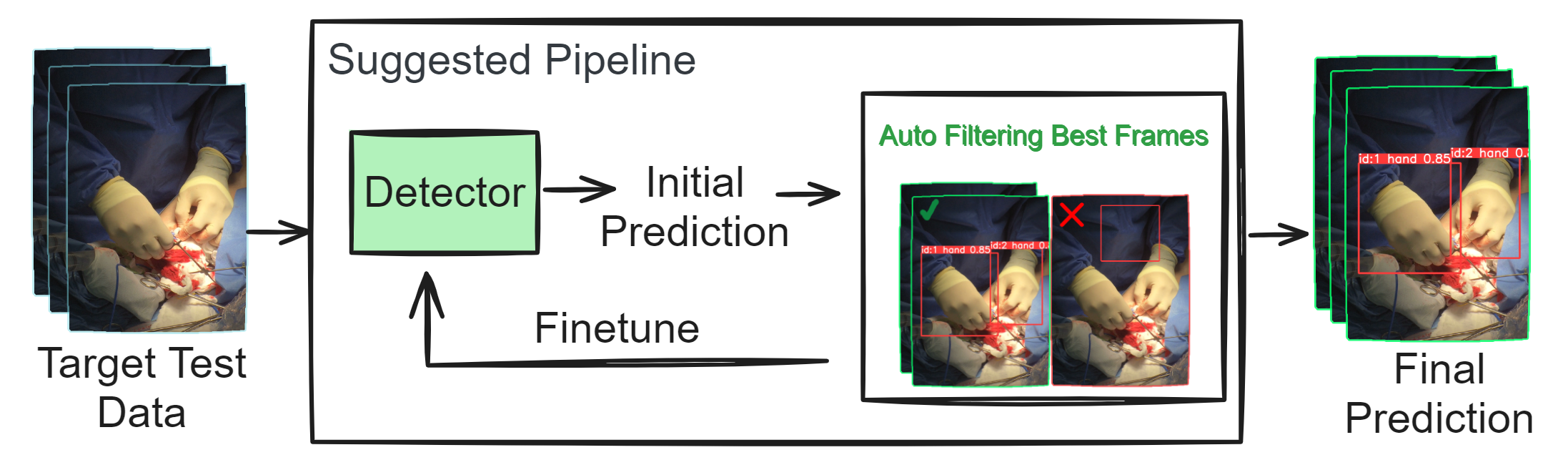}
    \caption{Our Semi-Supervised Domain Adaptation Self-Training Pipeline} \label{ilustratoin}
\end{figure}

The proposed methodology, encapsulated in the fine-tuned Domain Adaptation Pipeline, is a robust framework for addressing the challenges posed by environmental variations in object detection tasks (see Fig. \ref{ilustratoin}).

\subsection{Pseudo Labels Enhancement techniques}
\begin{figure}[!htb]
    \centering
    \includegraphics[scale=0.25]{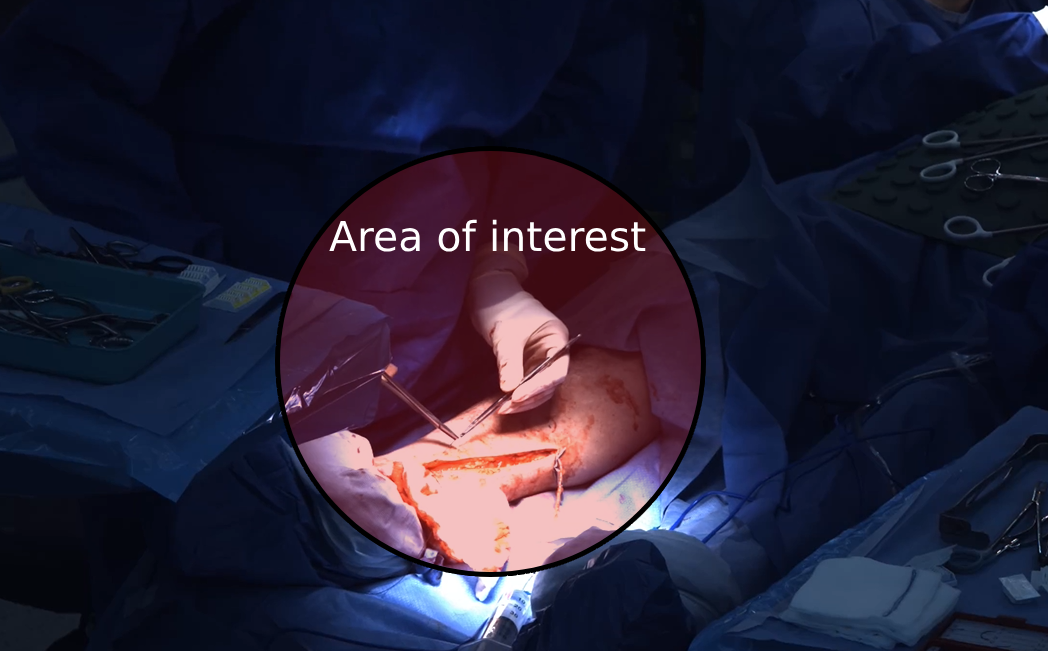}
   \caption{\textbf{Spacial Filtering} We filter out bounding boxes whose centers are out of the area of interest} \label{area_of_interest}
\end{figure}

We leverage the fact that our data source is video to filter the best frames automatically. Two methods were evaluated:

\noindent \textbf{Spacial Filtering}
Over time, the hands tend to remain within a limited area. Our method leverages this observation by defining an area of interest. Specifically, for every 10 consecutive frames, we calculate the average center of all bounding box centers, establishing the center of the area of interest. The radius of this area is set as a hyperparameter. During analysis, if the predicted center of a bounding box falls outside this defined circle, the corresponding prediction is filtered out. (see Fig. \ref{area_of_interest})

\noindent \textbf{Tracking filtering}
Utilizing the BotSort\cite{aharon2022bot} tracking algorithm, each prediction is assigned a unique identifier. Subsequently,  predictions associated with identifiers that have fewer than five instances are excluded. This criterion helps identify and eliminate predictions for objects that only appear for a brief duration.

\subsection{Implementation}
For detection training, we used Pytorch (ver 2.0) YOLOV8 \cite{Jocher_Ultralytics_YOLO_2023} implementation on Python version 3.10.9. Training was executed using NVIDIA RTX A6000 GPU. Initial experiments were performed to ascertain the optimal hyperparameters. Consistency was ensured by employing identical parameters across all subsequent experiments.

\section{Experiments and Results}
\subsection{Yolo generalization comparison}
This experiment assessed the model's ability to generalize across datasets from different distributions. For the initial training phase of the hand detection model, we partitioned all datasets into training and validation subsets, allocating $80\%$ and $20\%$, respectively. Subsequently, in every experiment, a detection model was trained using training and validation sets from the same dataset. The trained hands detector was then evaluated on the validation subsets of the remaining datasets (Table 
\ref{tab:yolo_generalization} ).

Our results indicate that each model achieved satisfactory performance when evaluated on validation data from the same distribution; however, their performance dramatically deteriorated when evaluated on different validation subsets.

% \begin{table}[h]
%     \centering
%     \label{tab:generalization}
%     \resizebox{\columnwidth}{!}{
%         \begin{tabular}{|c|c|c|c|c|c|c|c|c|c|}
%             \hline
%             \multirow{2}{*}{\textbf{Train\textbackslash Validation}} & \multicolumn{3}{c|}{\textbf{EgoHands}} & \multicolumn{3}{c|}{\textbf{OneHand10K}} & \multicolumn{3}{c|}{\textbf{HaDR}} \\ \cline{}
%             & \textbf{Precision} & \textbf{Recall} & \textbf{mAP50} & \textbf{Precision} & \textbf{Recall} & \textbf{mAP50} & \textbf{Precision} & \textbf{Recall} & \textbf{mAP50} \\ \hline
%             EgoHands & 0.983 & 0.947 & 0.991 & 0.525 & 0.471 & 0.402 & 0.343 & 0.137 & 0.130 \\ \hline
%             OneHand10K & 0.748 & 0.545 & 0.618 & 0.880 & 0.878 & 0.863 & 0.800 & 0.515 & 0.600 \\ \hline
%             HaDR & 0.336 & 0.222 & 0.175 & 0.522 & 0.412 & 0.335 & 0.997 & 0.992 & 0.995 \\ \hline
%         \end{tabular}
%     }
%     \caption{Yolo generalization comparison}
% \end{table}

\begin{table}[b]
    \centering
    \caption{Yolo generalization comparison}\label{tab:yolo_generalization}
    \begin{tabular}{|c|ccc|ccc|ccc|}
        \hline
        \multirow{2}{*}{Train\textbackslash{}Validation} & \multicolumn{3}{c|}{EgoHands} & \multicolumn{3}{c|}{OneHand10K} & \multicolumn{3}{c|}{HaDR} \\ 
        % \cline{} 
         & \multicolumn{1}{c|}{Prec} & \multicolumn{1}{c|}{Rec} & mAP & 
           \multicolumn{1}{c|}{Prec} & \multicolumn{1}{c|}{Rec} & mAP & 
           \multicolumn{1}{c|}{Prec} & \multicolumn{1}{c|}{Rec} & mAP \\ 
        \hline
        EgoHands & 
          \multicolumn{1}{c|}{0.983} & \multicolumn{1}{c|}{0.947} & 0.991 &
          \multicolumn{1}{c|}{0.525} & \multicolumn{1}{c|}{0.471} & 0.402 &
          \multicolumn{1}{c|}{0.343} & \multicolumn{1}{c|}{0.137} & 0.130 \\ 
        \hline
        OneHand10K &
          \multicolumn{1}{c|}{0.748} & \multicolumn{1}{c|}{0.545} & 0.618 &
          \multicolumn{1}{c|}{0.880} & \multicolumn{1}{c|}{0.878} & 0.863 &
          \multicolumn{1}{c|}{0.800} & \multicolumn{1}{c|}{0.515} & 0.600 \\ 
        \hline
        HaDR &
          \multicolumn{1}{c|}{0.336} & \multicolumn{1}{c|}{0.222} & 0.175 &
          \multicolumn{1}{c|}{0.522} & \multicolumn{1}{c|}{0.412} & 0.335 &
          \multicolumn{1}{c|}{0.997} & \multicolumn{1}{c|}{0.992} & 0.995 \\ 
        \hline
    \end{tabular}
    \begin{minipage}{\textwidth}
        \vspace{0.1cm} % Adjust vertical space as needed
        
        \small *Prec, Rec and mAP are Precision, Recall and mAP50
    \end{minipage}
\end{table}

% \begin{table}[!ht]
%    \centering
%         \caption{Yolo generalization comparison}\label{tab:generalization}
%     \begin{tabular}{|p{2.4cm}|p{2.4cm}|p{2.4cm}|p{2.4cm}|}
%     \hline
%         Train$\backslash$Validation & EgoHands  & OneHand10K & HaDR  \\ \hline
%         EgoHands  & precision 0.983
% recall 0.974
% mAP50 \hfill 0.991 & precision 0.525
% recall 0.471
% mAP50 \hfill 0.402 & precision 0.343
% recall 0.137
% mAP50 \hfill 0.130 \\ \hline
%         OneHand10K   & precision 0.748
% recall 0.545
% mAP50 \hfill 0.618 & precision 0.880
% recall 0.878
% mAP50 \hfill 0.863 & precision 0.800
% recall 0.515
% mAP50 \hfill 0.600 \\ \hline
%         HaDr  & precision 0.336
% recall 0.222
% mAP50 \hfill 0.175 & precision 0.522
% recall 0.412
% mAP50 \hfill 0.335 & precision 0.997
% recall 0.992
% mAP50 \hfill 0.995 \\ \hline
%     \end{tabular}
% \end{table}

\subsection{Creation of the Pre-trained Model}
To create a robust pre-trained model, we first checked the impact of each dataset with and without gloves. Then, we trained a YOLO model using all datasets (EgoHands, EgoHands + Artificial Gloves, OneHand10K, OneHand10K + Artificial Gloves, HaDR, Assembly101 sample + Artificial Gloves).   Subsequently, we performed validation analyses using the ERS and SVGH test cases. The model trained using the combination dataset presents the best results (see Table \ref{tab1}). 

\begin{table}[!htb]
\caption{Evaluation of the pre-trained models}\label{tab1}
\centering
\begin{tabular}{|l|l|l|l|l|l|l|}
\hline
\multicolumn{1}{|c|}{ } & \multicolumn{3}{c|}{ERS} & \multicolumn{3}{c|}{SVGH} \\
\hline
Train dataset &  Precision & Recall & mAP50 &  Precision & Recall & mAP50\\
\hline
Egohands & 0.775 & 0.447 & 0.534 &0.502 &0.358 &0.315 \\
Egohands + Artificial Gloves & 0.829 & 0.434 & 0.536 &0.527 &0.469 &0.422\\
OneHand10K & 0.759 & 0.406 & 0.524 &0.685 &0.515 &0.576\\
OneHand10K + Artificial Gloves & 0.813 & 0.468 & 0.560 &0.763 &0.545 &0.639\\
HaDR & 0.604 & 0.417 & 0.446 &0.447 &0.441 &0.369\\
{\bfseries Datasets combination} & {\bfseries 0.868} & {\bfseries 0.653} & {\bfseries 0.751} &{\bfseries 0.789} &{\bfseries 0.596} &{\bfseries 0.662}\\
\hline
\end{tabular}
\end{table}

\subsection{Fine Tuning using the Proposed Pipeline}
\begin{figure}[!htb]
    \centering
    \includegraphics[scale=0.22]{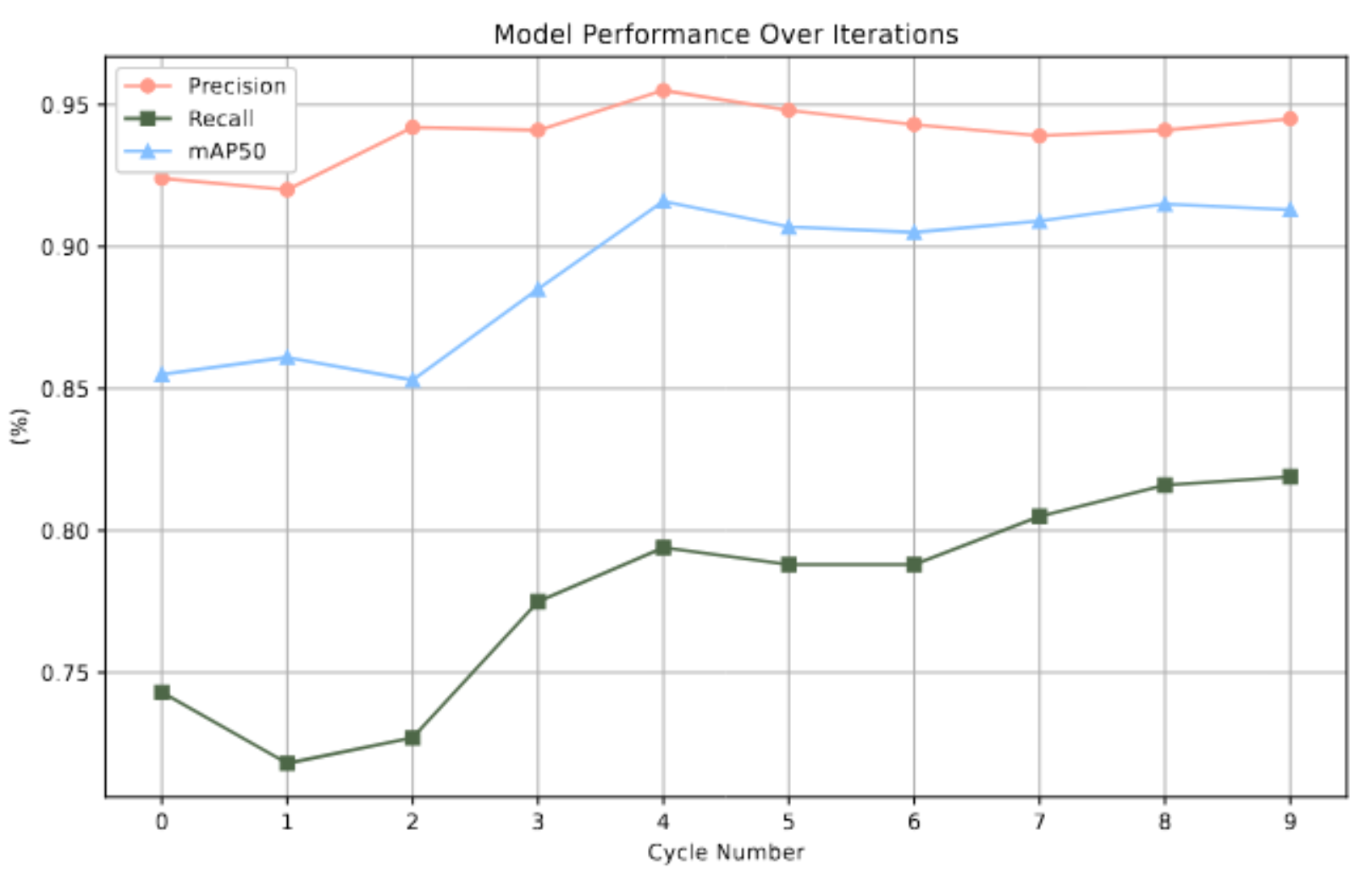}
    \caption{model performance during refinement pipeline on ERS} \label{fig iterative}
\end{figure}

In this phase, we fine-tuned the pre-trained model for each test case (ERS and SVGH), using data from that specific dataset. Several degrees of freedom exist, including the number of epochs for a fine-tuning session, the number of iterations within the pipeline, and the techniques for filtering suitable frames. Throughout all experiments, the fine-tuning session consisted of 5 epochs, while different runs employed varying numbers of refinement
cycles and distinct filtering techniques before training. (see Table \ref{Pipeline}). 
We can observe that the performance improves throughout the pipeline cycles, as illustrated in the figure with the ERS example (see Fig. \ref{fig iterative}).

\begin{table}
\caption{Fine Tuning Using The Proposed Pipeline model }\label{Pipeline}
\begin{tabular}{|p{3cm}|p{1.8cm}|l|l|l|l|l|l|}
\hline
\multicolumn{2}{|c|}{ } & \multicolumn{3}{c|}{ERS} & \multicolumn{3}{c|}{SVGH} \\
\hline
Filtering Heuristic\newline Applied & Refinements \newline Cycles   &  Precision & Recall & mAP50 &  Precision & Recall & mAP50\\
\hline
No Fine Tuning& 1 & 0.906 & 0.726 &0.818 &0.797 &0.525 &0.669 \\
 Area Of Interest&1& 0.917 & 0.726 & 0.840 &0.822 &0.672 &0.789 \\
 Tracking&1 & 0.924 & 0.743 & 0.855 &0.888 &0.709 &0.852 \\
{\bfseries  Tracking} &10& {\bfseries 0.945} &{\bfseries 0.819} & {\bfseries 0.913} &{\bfseries 0.940} &{\bfseries 0.850} &{\bfseries 0.912} \\
\hline
\end{tabular}
\end{table}

\subsection{Comparison To State-Of-The-Art Hand Models}
\begin{figure}[!htb]
    
    \includegraphics[width=\textwidth]{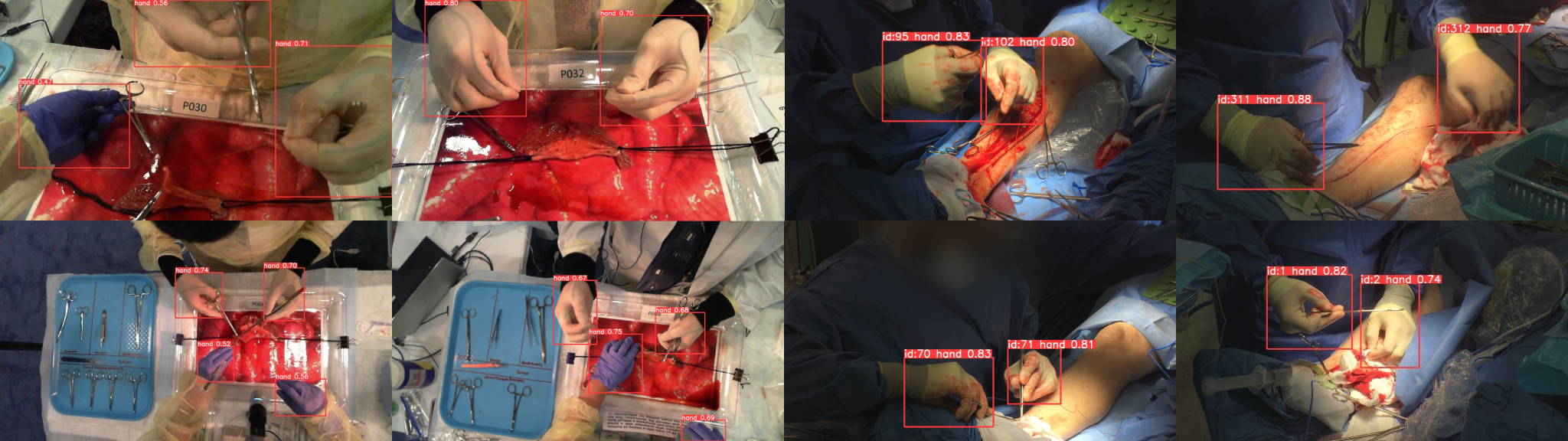}
    \caption{Rohan Hands detection on the validation sets of ERS (left) and SVGH (right) } 
    \label{fig validation_sets}
\end{figure}

Finally, we compared our model with hand-detection components of state-of-the-art hand models focusing on their effectiveness across two datasets, ERS and SVGH, using metrics such as Precision, Recall, and mAP50. our model RoHan with its best configuration (10 refinements
cycles) achieves the highest Precision (0.945 for ERS and 0.940 for SVGH) and mAP50 (0.913 for ERS and 0.912 for SVGH) across both datasets. This indicates RoHan's superior ability to accurately detect hand instances with minimal false positives (see table \ref{tab:Sota_comparison}). Figure \ref{fig validation_sets} depicts detection examples on both datasets.

\begin{table}
\caption{Comparison to Hand detection components of SOTA Hand models}\label{tab:Sota_comparison}
\begin{tabular}{|p{3cm}|l|l|l|l|l|l|}
\hline
\multicolumn{1}{|c|}{ } & \multicolumn{3}{c|}{ERS} & \multicolumn{3}{c|}{SVGH} \\
\hline
Method Applied & Precision & Recall & mAP50  &  Precision & Recall & mAP50 \\
\hline
Robust-CSD\cite{zhang2020} & 0.559 & 0.664& 0.591 &0.477 &0.608 & 0.572\\
\hline
HaMeR\cite{pavlakos2024} & 0.829 & 0.701 & 0.673 &0.855 &{\bfseries 0.983} & 0.855 \\
\hline
WiLoR\cite{potamias2024} & 0.926 & 0.709 & 0.750 &0.927 & 0.863 & 0.846 \\
\hline
{\bfseries  RoHan(ours)} & {\bfseries 0.945} &{\bfseries 0.819}& {\bfseries 0.913} &{\bfseries 0.940} &0.850 & {\bfseries 0.912} \\
\hline
\end{tabular}
\end{table}

% \begin{figure}[!htb]
%     \centering
%     \includegraphics[scale=0.33]{Figures/leg_detection.png}
%     \caption{Robustness Experiments on Leg Vein Removal Procedure} 
%     \label{fig Vein}
% \end{figure}

\section{Discussion and Conclusion}
Our study introduces an advanced pre-training strategy to develop a hand detection model that can be easily adapted to various surgical environments and camera conditions. Our method eliminates the need to label new data by introducing domain-specific augmentation and iterative self-training refinement processes; additionally, we offer benchmark datasets for future research on hand detection.

Hand detection is crucial for advanced applications like 3D hand reconstruction, hand-object completion, and segmentation, all vital for skill assessment. This step ensures precise hand localization within images or videos, facilitating further analyses such as pose estimation and interaction analysis. Accurate hand detection is critical in 3D hand reconstruction to create exact models of the hand's structure and movement. Integrating our robust detection model can enhance state-of-the-art computer vision algorithms. Additionally, the concept of our Domain Adaptation iterative pipeline can improve the predictive capabilities of various models.

Hand detection is crucial in many fields beyond surgery. Despite the general purpose of existing datasets, a significant domain gap limits model transferability. To overcome this, our pre-training combines all available datasets to leverage their diversity. Additionally, to cater to the unique requirements of OR data, we introduce domain-specific augmentations: Artificial Gloves.

Additionally, we introduce an iterative prediction refinement process that minimizes the domain gap. This involves generating pseudo labels with the initial model, which are then used to refine the model. The effectiveness of this refinement depends on the quality of the pseudo labels. Consequently, we leverage the temporal characteristics of video data to eliminate potentially low-quality pseudo labels. The impact of the refinement process and the use of temporal information to enhance the pseudo label is not limited to our work. For example, even when labeling was performed, the domain may shift over time (camera position might change, lighting replaced, etc.). Our suggested refinement approach may used to retrain the model. Furthermore, this method is not limited to hands, for example it may be used for tool detection.

Our research has several limitations. Primarily, the effectiveness of the refinement pipeline relies on the quality of the initial pseudo-labels. This was addressed in the data enhancement step. However, if the dataset doesn't include video data, different enhancement methods may be required.

Our contribution of benchmark datasets aims to catalyze further research in this domain, offering a resource for developing and testing new hand detection models. As the field evolves, we anticipate our approach will lay the groundwork for more sophisticated computer vision applications in medicine and beyond.

\bibliography{sn-bibliography}% common bib file


\begin{thebibliography}{18}
\ifx \bisbn   \undefined \def \bisbn  #1{ISBN #1}\fi
\ifx \binits  \undefined \def \binits#1{#1}\fi
\ifx \bauthor  \undefined \def \bauthor#1{#1}\fi
\ifx \batitle  \undefined \def \batitle#1{#1}\fi
\ifx \bjtitle  \undefined \def \bjtitle#1{#1}\fi
\ifx \bvolume  \undefined \def \bvolume#1{\textbf{#1}}\fi
\ifx \byear  \undefined \def \byear#1{#1}\fi
\ifx \bissue  \undefined \def \bissue#1{#1}\fi
\ifx \bfpage  \undefined \def \bfpage#1{#1}\fi
\ifx \blpage  \undefined \def \blpage #1{#1}\fi
\ifx \burl  \undefined \def \burl#1{\textsf{#1}}\fi
\ifx \doiurl  \undefined \def \doiurl#1{\url{https://doi.org/#1}}\fi
\ifx \betal  \undefined \def \betal{\textit{et al.}}\fi
\ifx \binstitute  \undefined \def \binstitute#1{#1}\fi
\ifx \binstitutionaled  \undefined \def \binstitutionaled#1{#1}\fi
\ifx \bctitle  \undefined \def \bctitle#1{#1}\fi
\ifx \beditor  \undefined \def \beditor#1{#1}\fi
\ifx \bpublisher  \undefined \def \bpublisher#1{#1}\fi
\ifx \bbtitle  \undefined \def \bbtitle#1{#1}\fi
\ifx \bedition  \undefined \def \bedition#1{#1}\fi
\ifx \bseriesno  \undefined \def \bseriesno#1{#1}\fi
\ifx \blocation  \undefined \def \blocation#1{#1}\fi
\ifx \bsertitle  \undefined \def \bsertitle#1{#1}\fi
\ifx \bsnm \undefined \def \bsnm#1{#1}\fi
\ifx \bsuffix \undefined \def \bsuffix#1{#1}\fi
\ifx \bparticle \undefined \def \bparticle#1{#1}\fi
\ifx \barticle \undefined \def \barticle#1{#1}\fi
\bibcommenthead
\ifx \bconfdate \undefined \def \bconfdate #1{#1}\fi
\ifx \botherref \undefined \def \botherref #1{#1}\fi
\ifx \url \undefined \def \url#1{\textsf{#1}}\fi
\ifx \bchapter \undefined \def \bchapter#1{#1}\fi
\ifx \bbook \undefined \def \bbook#1{#1}\fi
\ifx \bcomment \undefined \def \bcomment#1{#1}\fi
\ifx \oauthor \undefined \def \oauthor#1{#1}\fi
\ifx \citeauthoryear \undefined \def \citeauthoryear#1{#1}\fi
\ifx \endbibitem  \undefined \def \endbibitem {}\fi
\ifx \bconflocation  \undefined \def \bconflocation#1{#1}\fi
\ifx \arxivurl  \undefined \def \arxivurl#1{\textsf{#1}}\fi
\csname PreBibitemsHook\endcsname

%%% 1
\bibitem[\protect\citeauthoryear{Kim et~al.}{2015}]{kim2015}
\begin{barticle}
\bauthor{\bsnm{Kim}, \binits{F.J.}},
\bauthor{\bsnm{Silva}, \binits{R.D.}},
\bauthor{\bsnm{Gustafson}, \binits{D.}},
\bauthor{\bsnm{Nogueira}, \binits{L.}},
\bauthor{\bsnm{Harlin}, \binits{T.}},
\bauthor{\bsnm{Paul}, \binits{D.L.}}:
\batitle{Current issues in patient safety in surgery: a review}.
\bjtitle{Patient safety in surgery}
\bvolume{9},
\bfpage{1}--\blpage{9}
(\byear{2015})
\end{barticle}
\endbibitem

%%% 2
\bibitem[\protect\citeauthoryear{Huang et~al.}{2011}]{huang2011}
\begin{barticle}
\bauthor{\bsnm{Huang}, \binits{F.C.}},
\bauthor{\bsnm{Mussa-Ivaldi}, \binits{F.A.}},
\bauthor{\bsnm{Pugh}, \binits{C.M.}},
\bauthor{\bsnm{Patton}, \binits{J.L.}}:
\batitle{Learning kinematic constraints in laparoscopic surgery}.
\bjtitle{IEEE transactions on haptics}
\bvolume{5}(\bissue{4}),
\bfpage{356}--\blpage{364}
(\byear{2011})
\end{barticle}
\endbibitem

%%% 3
\bibitem[\protect\citeauthoryear{Ward et~al.}{2021}]{ward2021}
\begin{barticle}
\bauthor{\bsnm{Ward}, \binits{T.M.}},
\bauthor{\bsnm{Mascagni}, \binits{P.}},
\bauthor{\bsnm{Ban}, \binits{Y.}},
\bauthor{\bsnm{Rosman}, \binits{G.}},
\bauthor{\bsnm{Padoy}, \binits{N.}},
\bauthor{\bsnm{Meireles}, \binits{O.}},
\bauthor{\bsnm{Hashimoto}, \binits{D.A.}}:
\batitle{Computer vision in surgery}.
\bjtitle{Surgery}
\bvolume{169}(\bissue{5}),
\bfpage{1253}--\blpage{1256}
(\byear{2021})
\end{barticle}
\endbibitem

%%% 4
\bibitem[\protect\citeauthoryear{Zhang et~al.}{2020}]{zhang2020}
\begin{bchapter}
\bauthor{\bsnm{Zhang}, \binits{M.}},
\bauthor{\bsnm{Cheng}, \binits{X.}},
\bauthor{\bsnm{Copeland}, \binits{D.}},
\bauthor{\bsnm{Desai}, \binits{A.}},
\bauthor{\bsnm{Guan}, \binits{M.Y.}},
\bauthor{\bsnm{Brat}, \binits{G.A.}},
\bauthor{\bsnm{Yeung}, \binits{S.}}:
\bctitle{Using computer vision to automate hand detection and tracking of surgeon movements in videos of open surgery}.
In: \bbtitle{AMIA Annual Symposium Proceedings},
vol. \bseriesno{2020},
p. \bfpage{1373}
(\byear{2020}).
\bcomment{American Medical Informatics Association}
\end{bchapter}
\endbibitem

%%% 5
\bibitem[\protect\citeauthoryear{Vaid et~al.}{2023}]{vaid2023}
\begin{bchapter}
\bauthor{\bsnm{Vaid}, \binits{P.}},
\bauthor{\bsnm{Yeung}, \binits{S.}},
\bauthor{\bsnm{Rau}, \binits{A.}}:
\bctitle{Robust semi-supervised detection of hands in diverse open surgery environments}.
In: \bbtitle{Machine Learning for Healthcare Conference},
pp. \bfpage{736}--\blpage{753}
(\byear{2023}).
\bcomment{PMLR}
\end{bchapter}
\endbibitem

%%% 6
\bibitem[\protect\citeauthoryear{Chen et~al.}{2019}]{chen2019}
\begin{bchapter}
\bauthor{\bsnm{Chen}, \binits{S.-T.}},
\bauthor{\bsnm{Cornelius}, \binits{C.}},
\bauthor{\bsnm{Martin}, \binits{J.}},
\bauthor{\bsnm{Chau}, \binits{D.H.}}:
\bctitle{Shapeshifter: Robust physical adversarial attack on faster r-cnn object detector}.
In: \bbtitle{Machine Learning and Knowledge Discovery in Databases: European Conference, ECML PKDD 2018, Dublin, Ireland, September 10--14, 2018, Proceedings, Part I 18},
pp. \bfpage{52}--\blpage{68}
(\byear{2019}).
\bcomment{Springer}
\end{bchapter}
\endbibitem

%%% 7
\bibitem[\protect\citeauthoryear{Basiev Kristina ;~Goldbraikh}{2022}]{basiev2022open}
\begin{barticle}
\bauthor{\bsnm{Basiev Kristina ;~Goldbraikh}, \binits{A.e.a.}}:
\batitle{Open surgery tool classification and hand utilization using a multi-camera system}.
\bjtitle{International journal of computer assisted radiology and surgery}
\bvolume{17},
\bfpage{1497}--\blpage{1505}
(\byear{2022})
\end{barticle}
\endbibitem

%%% 8
\bibitem[\protect\citeauthoryear{Jiang et~al.}{2022}]{jiang2022}
\begin{barticle}
\bauthor{\bsnm{Jiang}, \binits{P.}},
\bauthor{\bsnm{Ergu}, \binits{D.}},
\bauthor{\bsnm{Liu}, \binits{F.}},
\bauthor{\bsnm{Cai}, \binits{Y.}},
\bauthor{\bsnm{Ma}, \binits{B.}}:
\batitle{A review of yolo algorithm developments}.
\bjtitle{Procedia Computer Science}
\bvolume{199},
\bfpage{1066}--\blpage{1073}
(\byear{2022})
\end{barticle}
\endbibitem

%%% 9
\bibitem[\protect\citeauthoryear{Goodman et~al.}{2021}]{goodman2021}
\begin{botherref}
\oauthor{\bsnm{Goodman}, \binits{E.D.}},
\oauthor{\bsnm{Patel}, \binits{K.K.}},
\oauthor{\bsnm{Zhang}, \binits{Y.}},
\oauthor{\bsnm{Locke}, \binits{W.}},
\oauthor{\bsnm{Kennedy}, \binits{C.J.}},
\oauthor{\bsnm{Mehrotra}, \binits{R.}},
\oauthor{\bsnm{Ren}, \binits{S.}},
\oauthor{\bsnm{Guan}, \binits{M.Y.}},
\oauthor{\bsnm{Downing}, \binits{M.}},
\oauthor{\bsnm{Chen}, \binits{H.W.}}, et al.:
A real-time spatiotemporal ai model analyzes skill in open surgical videos.
arXiv preprint arXiv:2112.07219
(2021)
\end{botherref}
\endbibitem

%%% 10
\bibitem[\protect\citeauthoryear{Pavlakos et~al.}{2024}]{pavlakos2024}
\begin{botherref}
\oauthor{\bsnm{Pavlakos}, \binits{G.}},
\oauthor{\bsnm{Shan}, \binits{D.}},
\oauthor{\bsnm{Radosavovic}, \binits{I.}},
\oauthor{\bsnm{Kanazawa}, \binits{A.}},
\oauthor{\bsnm{Fouhey}, \binits{D.}},
\oauthor{\bsnm{Malik}, \binits{J.}}:
Reconstructing hands in 3d with transformers.
CVPR
(2024)
\end{botherref}
\endbibitem

%%% 11
\bibitem[\protect\citeauthoryear{Potamias et~al.}{2024}]{potamias2024}
\begin{botherref}
\oauthor{\bsnm{Potamias}, \binits{R.A.}},
\oauthor{\bsnm{Zhang}, \binits{J.}},
\oauthor{\bsnm{Deng}, \binits{J.}},
\oauthor{\bsnm{Zafeiriou}, \binits{S.}}:
WiLoR: End-to-end 3D Hand Localization and Reconstruction in-the-wild
(2024)
\end{botherref}
\endbibitem

%%% 12
\bibitem[\protect\citeauthoryear{Bambach et~al.}{2015}]{bambach2015}
\begin{bchapter}
\bauthor{\bsnm{Bambach}, \binits{S.}},
\bauthor{\bsnm{Lee}, \binits{S.}},
\bauthor{\bsnm{Crandall}, \binits{D.J.}},
\bauthor{\bsnm{Yu}, \binits{C.}}:
\bctitle{Lending a hand: Detecting hands and recognizing activities in complex egocentric interactions}.
In: \bbtitle{Proceedings of the IEEE International Conference on Computer Vision},
pp. \bfpage{1949}--\blpage{1957}
(\byear{2015})
\end{bchapter}
\endbibitem

%%% 13
\bibitem[\protect\citeauthoryear{Wang et~al.}{2018}]{wang2018mask}
\begin{barticle}
\bauthor{\bsnm{Wang}, \binits{Y.}},
\bauthor{\bsnm{Peng}, \binits{C.}},
\bauthor{\bsnm{Liu}, \binits{Y.}}:
\batitle{Mask-pose cascaded cnn for 2d hand pose estimation from single color image}.
\bjtitle{IEEE Transactions on Circuits and Systems for Video Technology}
\bvolume{29}(\bissue{11}),
\bfpage{3258}--\blpage{3268}
(\byear{2018})
\end{barticle}
\endbibitem

%%% 14
\bibitem[\protect\citeauthoryear{Grushko et~al.}{2023}]{grushko2023}
\begin{botherref}
\oauthor{\bsnm{Grushko}, \binits{S.}},
\oauthor{\bsnm{Vysock{\`y}}, \binits{A.}},
\oauthor{\bsnm{Chlebek}, \binits{J.}},
\oauthor{\bsnm{Prokop}, \binits{P.}}:
Hadr: Applying domain randomization for generating synthetic multimodal dataset for hand instance segmentation in cluttered industrial environments.
arXiv preprint arXiv:2304.05826
(2023)
\end{botherref}
\endbibitem

%%% 15
\bibitem[\protect\citeauthoryear{Sener et~al.}{2022}]{sener2022assembly101}
\begin{bchapter}
\bauthor{\bsnm{Sener}, \binits{F.}},
\bauthor{\bsnm{Chatterjee}, \binits{D.}},
\bauthor{\bsnm{Shelepov}, \binits{D.}},
\bauthor{\bsnm{He}, \binits{K.}},
\bauthor{\bsnm{Singhania}, \binits{D.}},
\bauthor{\bsnm{Wang}, \binits{R.}},
\bauthor{\bsnm{Yao}, \binits{A.}}:
\bctitle{Assembly101: A large-scale multi-view video dataset for understanding procedural activities}.
In: \bbtitle{Proceedings of the IEEE/CVF Conference on Computer Vision and Pattern Recognition},
pp. \bfpage{21096}--\blpage{21106}
(\byear{2022})
\end{bchapter}
\endbibitem

%%% 16
\bibitem[\protect\citeauthoryear{Moon}{2023}]{moon2023bringing}
\begin{bchapter}
\bauthor{\bsnm{Moon}, \binits{G.}}:
\bctitle{Bringing inputs to shared domains for 3d interacting hands recovery in the wild}.
In: \bbtitle{Proceedings of the IEEE/CVF Conference on Computer Vision and Pattern Recognition},
pp. \bfpage{17028}--\blpage{17037}
(\byear{2023})
\end{bchapter}
\endbibitem

%%% 17
\bibitem[\protect\citeauthoryear{Aharon et~al.}{2022}]{aharon2022bot}
\begin{botherref}
\oauthor{\bsnm{Aharon}, \binits{N.}},
\oauthor{\bsnm{Orfaig}, \binits{R.}},
\oauthor{\bsnm{Bobrovsky}, \binits{B.-Z.}}:
Bot-sort: Robust associations multi-pedestrian tracking.
arXiv preprint arXiv:2206.14651
(2022)
\end{botherref}
\endbibitem

%%% 18
\bibitem[\protect\citeauthoryear{Jocher et~al.}{2023}]{Jocher_Ultralytics_YOLO_2023}
\begin{botherref}
\oauthor{\bsnm{Jocher}, \binits{G.}},
\oauthor{\bsnm{Chaurasia}, \binits{A.}},
\oauthor{\bsnm{Qiu}, \binits{J.}}:
{Ultralytics YOLO}.
\url{https://github.com/ultralytics/ultralytics}
\end{botherref}
\endbibitem

\end{thebibliography}

\end{document}